\documentclass{ngsm2024}

\hypersetup{}
\usepackage{siunitx}
\usepackage[symbol]{footmisc}
\usepackage{footnote}
\usepackage{tikz}
\usetikzlibrary {arrows.meta}
\tikzset{triangle/.style={
    arrows={-Stealth[inset=0pt, angle=90:3pt]},
    fill=gray,
    draw=gray,
}}
\tikzset{horizontal/.style={
    fill=gray!20,
    rounded corners=0.1cm,
    minimum width=3cm,
    minimum height=0.5cm,
}}
\tikzset{vertical/.style={
    fill=gray!20,
    rounded corners=0.1cm,
    minimum width=0.5cm,
    minimum height=1cm,
}}
\tikzset{neural/.style={
    fill=blue!20,
    rounded corners=0.1cm,
    minimum width=3cm,
    minimum height=1cm,
}}
\tikzset{blank/.style={
    fill=none,
    rounded corners=0.1cm,
    minimum width=3cm,
    minimum height=0.48cm,
    dashed,
    draw=black,
}}
\usepackage{parskip}
\def\diff#1{\operatorname{d}\!{#1}}

\DeclareMathOperator*{\argmin}{arg\,min}
\DeclareMathOperator*{\argmax}{arg\,max}
\SetAlCapHSkip{0em}
\SetKw{KwInput}{input:}
\SetKw{KwReturn}{return:}
\usepackage{booktabs}
\usepackage{pifont}
\newcommand{\cmark}{{\color{Green}\ding{51}}}
\newcommand{\xmark}{{\color{Red}\ding{55}}}

\title[%
    Parallelizing Autoregressive Generation with Variational State Space Models
]{%
    Parallelizing Autoregressive Generation with \\ Variational State Space Models
}

\optauthor{%
    \Name{Gaspard Lambrechts}\footnotemark[1] \Email{gaspard.lambrechts@uliege.be} \\
    \Name{Yann Claes}\footnotemark[1] \Email{y.claes@uliege.be} \\
    \Name{Pierre Geurts} \Email{p.geurts@uliege.be} \\
    \Name{Damien Ernst} \Email{dernst@uliege.be} \\
\addr Montefiore Institute, University of Liège}

\begin{document}

\maketitle

\setcounter{footnote}{1}
\footnotetext[1]{Equal contributions.}

\begin{abstract}%
    Attention-based models such as Transformers and recurrent models like state space models (SSMs) have emerged as successful methods for autoregressive sequence modeling.
    Although both enable parallel training, none enable parallel generation due to their autoregressiveness.
    We propose the variational SSM (VSSM), a variational autoencoder (VAE) where both the encoder and decoder are SSMs.
    Since sampling the latent variables and decoding them with the SSM can be parallelized, both training and generation can be conducted in parallel.
    Moreover, the decoder recurrence allows generation to be resumed without reprocessing the whole sequence.
    Finally, we propose the autoregressive VSSM that can be conditioned on a partial realization of the sequence, as is common in language generation tasks.
    Interestingly, the autoregressive VSSM still enables parallel generation.
    We highlight on toy problems (MNIST, CIFAR) the empirical gains in speed-up and show that it competes with traditional models in terms of generation quality (Transformer, Mamba SSM).
\end{abstract}


\section{Introduction} \label{sec:introduction}
Sequence modeling tasks, namely time-series forecasting and text generation, have gained in popularity and various types of architectures were designed to tackle such problems.
Transformers were proven effective \citep{vaswani2017attention, radford2019language}, yet they nonetheless reprocess the complete sequence at each timestep, making generation less efficient.
Recurrent neural networks (RNNs) \citep{graves2013generating, cho2014learning} update a hidden state based on new inputs at each timestep, enabling efficient generation.
SSMs \citep{gupta2022diagonal, gu2022parameterization, smith2023simplified, gu2023mamba}, a recently introduced class of RNNs, enable parallel training thanks to their linear recurrence.
Alternatively, several works adapt VAEs for sequential modeling.
Some architectures integrate Transformers \citep{liu2019transformer, jiang2020transformer} and enable parallel training, although little work \citep{fang2021transformer} proposes models that can be conditioned on partial realizations (e.g., prompts).
Conversely, variational RNNs (VRNNs) \citep{chung2015recurrent} loose parallelizability by making the model both autoregressive and recurrent, allowing it to be conditioned on partial realizations and to resume generation.
However, all introduced autoregressive models perform generation sequentially, as they are explicitly conditioned on previously generated data.

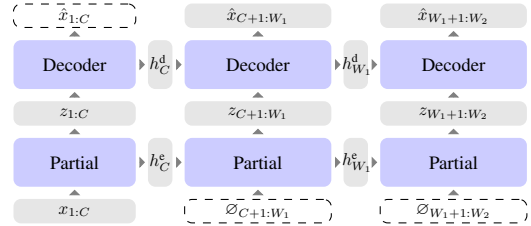
\begin{figure}[ht]
    \vspace{-1.0em}
    \centering
    \subfigure[\small Time complexities and parallelizability at training and sampling, and generation properties.]{%
        \scriptsize
        \label{tab:complexities}
        \setlength{\tabcolsep}{3pt}
        \begin{tabular}[t]{c|cccccc}
            \toprule
            \textbf{Model} & \textbf{Training} & \textbf{//} & \textbf{Sampling} & \textbf{//} & \textbf{Prompt} & \textbf{Resume} \\
            \midrule
            Transformer & $O(T^2)$ & \cmark & $O(T^2)$ & \xmark & \cmark & \xmark \\
            RNN & $O(T)$ & \xmark & $O(T)$ & \xmark & \cmark & \cmark \\
            SSM & $O(T)$ & \cmark & $O(T)$ & \xmark & \cmark & \cmark \\
            \midrule
            TVAE & $O(T^2)$ & \cmark & $O(T^2)$ & \cmark & \xmark/\cmark & \xmark \\
            VRNN & $O(T)$ & \xmark & $O(T)$ & \xmark & \cmark & \cmark \\
            VSSM & $O(T)$ & \cmark & $O(T)$ & \cmark & \cmark & \cmark \\
            \bottomrule
            \multicolumn{7}{c}{\vspace{-0.7em}}
        \end{tabular}
    }
    \hspace{0.4em}
    \subfigure[\small Parallel and recurrent sampling algorithm, given a contextual prompt $x_{1:C}$.]{%
        \label{fig:sampling}
        \begin{tikzpicture}[scale=0.65, transform shape]
            \node[horizontal, minimum width=2.5cm] (x0) at (0.25, 0) {};
                \node at (x0) {\small$x_{1:C}$};
            \node[blank] (x1) at (4, 0) {};
                \node at (x1) {\small$\varnothing_{C+1:W_1}$};
            \node[blank] (x2) at (8, 0) {};
                \node at (x2) {\small$\varnothing_{W_1+1:W_2}$};
            \draw[triangle] (0.25, 0.34) -- (0.25, 0.44);
            \draw[triangle] (4, 0.34) -- (4, 0.44);
            \draw[triangle] (8, 0.34) -- (8, 0.44);
            \node[neural, minimum width=2.5cm] (e0) at (0.25, 1) {Partial};
            \draw[triangle] (1.59, 1) -- (1.69, 1);
            \node[vertical] (he0) at (2, 1) {};
                \node at (he0) {\small$h_C^\text{e}$};
            \draw[triangle] (2.34, 1) -- (2.44, 1);
            \node[neural] (e1) at (4, 1) {Partial};
            \draw[triangle] (5.59, 1) -- (5.69, 1);
            \node[vertical] (he1) at (6, 1) {};
                \node[shift={(0.03, 0)}] at (he1) {\small$h_{W_1}^\text{e}$};
            \draw[triangle] (6.34, 1) -- (6.44, 1);
            \node[neural] (e2) at (8, 1) {Partial};
            \draw[triangle] (0.25, 1.59) -- (0.25, 1.69);
            \draw[triangle] (4, 1.59) -- (4, 1.69);
            \draw[triangle] (8, 1.59) -- (8, 1.69);
            \node[horizontal, minimum width=2.5cm] (z0) at (0.25, 2) {};
                \node at (z0) {\small$z_{1:C}$};
            \node[horizontal] (z1) at (4, 2) {};
                \node at (z1) {\small$z_{C+1:W_1}$};
            \node[horizontal] (z2) at (8, 2) {};
                \node at (z2) {\small$z_{W_1+1:W_2}$};
            \draw[triangle] (0.25, 2.34) -- (0.25, 2.44);
            \draw[triangle] (4, 2.34) -- (4, 2.44);
            \draw[triangle] (8, 2.34) -- (8, 2.44);
            \node[neural, minimum width=2.5cm] (d0) at (0.25, 3) {Decoder};
            \draw[triangle] (1.59, 3) -- (1.69, 3);
            \node[vertical] (hd0) at (2, 3) {};
                \node at (hd0) {\small$h_C^\text{d}$};
            \draw[triangle] (2.34, 3) -- (2.44, 3);
            \node[neural] (d1) at (4, 3) {Decoder};
            \draw[triangle] (5.59, 3) -- (5.69, 3);
            \node[vertical] (hd1) at (6, 3) {};
                \node[shift={(0.03, 0)}] at (hd1) {\small$h_{W_1}^\text{d}$};
            \draw[triangle] (6.34, 3) -- (6.44, 3);
            \node[neural] (d2) at (8, 3) {Decoder};
            \draw[triangle] (0.25, 3.59) -- (0.25, 3.69);
            \draw[triangle] (4, 3.59) -- (4, 3.69);
            \draw[triangle] (8, 3.59) -- (8, 3.69);
            \node[blank, minimum width=2.5cm] (r0) at (0.25, 4) {};
            \node at (r0) {\small$\hat{x}_{1:C}$};
            \node[horizontal] (r1) at (4, 4) {};
            \node at (r1) {\small$\hat{x}_{C+1:W_1}$};
            \node[horizontal] (r2) at (8, 4) {};
            \node at (r2) {\small$\hat{x}_{W_1+1:W_2}$};
        \end{tikzpicture}
    }
    \vspace{-0.5em}
    \label{fig:figure}
    \caption{Sequence models properties and VSSM sampling algorithm.}
    \vspace{-1.0em}
\end{figure}

Therefore, we propose the VSSM, a VAE whose encoder and decoder are SSMs.
Thanks to key architectural choices, both training and inference can be performed in parallel and linear time with respect to the sequence length, while still allowing generation to be to resumed without reprocessing the entire sequence.
In contrast, a VAE with Transformer encoder and decoder, which we call Transformer VAE (TVAE), would preserve parallel training and generation, but would not be resumable.
We then propose the autoregressive VSSM, that can be conditioned on partial realizations of the sequence and still generates in parallel.
The VSSM combines all advantages of previous models, as observed in \autoref{tab:complexities}, while producing results comparable to Transformers and SSMs on simple tasks (MNIST, CIFAR).
We highlight a recent work \citep{zhou2023deep} that proposes a similar architecture, yet their prior and generative models are explicitly autoregressive and do not exploit the parallelizability of SSMs.
Moreover, they only consider generation from sampled latents, while we also propose an approach to condition the model on partial realizations.
We do not consider diffusion models for sequences (e.g., \citep{gong2023diffuseq}), but note that they would not allow recurrent (i.e., resuming) generation.

\section{Background} \label{sec:background}

\subsection{Variational Autoencoders for Time Series} \label{subsec:vae}

We consider dynamical VAEs \citep{girin2021dynamical}, that model sequential data $x_{1:T}$ of length $T$ through $T$ latent variables $z_{1:T}$.
Given a target space $\mathcal{X}$, they define the joint distribution $p_\phi(x_{1:T}, z_{1:T})$ with,
\begin{itemize}
    \itemsep-0.3em
    \item A latent space $\mathcal{Z}$,
    \item A prior distribution $p_\phi(z_{1:T}) = \prod_{t=1}^T p_\phi(z_t | z_{1:t-1})$,
    \item A generative distribution $p_\phi(x_{1:T}|z_{1:T}) = \prod_{t=1}^T p_\phi(x_{t}|x_{1:t-1}, z_{1:T})$,
\end{itemize}
where $\phi$ denotes the parameters of these probability distributions.
Unfortunately, the likelihood of the data
$
    p_\phi(x_{1:T}) = \mathbb{E}_{p_\phi(z_{1:T})} p_\phi(x_{1:T}|z_{1:T})
$
under this model cannot be evaluated in practice.
Nevertheless, we can show that the log-likelihood is lower bounded by the evidence lower bound (ELBO), for any conditional probability distribution $q(z_{1:T}|x_{1:T})$,
\begin{align}
    \log p_{\phi}(x_{1:T})
    \geq \hspace{-0.5em}
        \mathop{\mathbb{E}}_{q(z_{1:T}|x_{1:T})} \log p_\phi(x_{1:T}|z_{1:T}) - \operatorname{KL}(q(z_{1:T}|x_{1:T}) \parallel p_\phi(z_{1:T}))
    =
        \operatorname{ELBO}_\phi(x_{1:T})
\end{align}
Moreover, the ELBO becomes tight when $q(z_{1:T}|x_{1:T})$ corresponds to the true posterior distribution $p_\phi(z_{1:T}|x_{1:T})$.
Thus, the generative model $p_\phi$ is usually jointly optimized with,
\begin{itemize}
    \item A posterior distribution $q_\psi(z_{1:T}|x_{1:T}) = \prod_{t=1}^T q_\psi(z_{t}|z_{1:t-1}, x_{1:T})$,
\end{itemize}
where $\psi$ denotes the parameters of this distribution.
These four components compose the dynamical VAE.
More details are provided in \autoref{app:derivations}.

\subsection{State Space Models} \label{subsec:ssm}

SSMs are linear and time-invariant dynamical systems that can be discretized into $h_{t} = A h_{t-1} + B u_{t}$, where $\zeta = (A,B)$ are learnable parameters.
Using the prefix-sum algorithm \citep{blelloch1990prefix}, we can parallelize the computation of the state sequence $h_{t} = \operatorname{SSM}_\zeta(u_{1:t})$ along all timesteps $t \in [1,T]$.
Furthermore, we can obtain effective sequence models of the form $y_{t} = f_\theta(u_{1:t})$ by stacking $L$ layers $i = \{1, \dots, L\}$ of interleaved SSMs and timestep-wise feedforward neural networks (FNNs),
\begin{equation}
h_{t}^{i}  = \operatorname{SSM}_{\zeta_{i}}(u_{1:t}^{i-1}), \quad\quad\quad
y_{t}^{i}  = \operatorname{FNN}_{\xi_{i}}(h_{t}^{i}),
\end{equation}
where $u_{t}^{i} = y_{t}^{i-1}$, $u_{t}^{0} = u_{t}$, $y_{t} = y_{t}^{L}$, and
$\theta = \cup_{i=1}^L(\zeta_{i}, \xi_{i})$ includes
all SSMs and FNNs parameters.
Indeed, it is believed that such stacking
of SSMs and timestep-wise FNNs is a universal approximator of
sufficiently regular non-linear sequence-to-sequence maps \citep{orvieto2023universality}.

\section{Method} \label{sec:method}

\subsection{Variational State Space Model} \label{subsec:vssm}

\setcounter{footnote}{0}
\renewcommand{\thefootnote}{\arabic{footnote}}

We introduce the VSSM as an instance of dynamical VAE, where we select, given a target space $\mathcal{X}$,
\begin{itemize}
    \itemsep-0.3em
    \item A discrete latent space $\mathcal{Z} = \{1, \dots, N\}^Z$ of $Z$ components of cardinality $N$ each,
    \item A uniform prior distribution $p_\phi(z_{1:T}) = \prod_{t=1}^T p_\phi(z_t | z_{1:t-1}) = \prod_{t=1}^T p_\phi(z_t) = \prod_{t=1}^T \frac{1}{N^Z}$,
    \item A generative distribution
    $p_\phi(x_{1:T}|z_{1:T})
    = \prod_{t=1}^T p_\phi(x_t|z_{1:t})
    = \prod_{t=1}^T\mathcal{P}(x_t|f^\text{dec}_\phi(z_{1:t}))$,
    where $\mathcal{P}(x_t|w_t)$\footnote{Gaussian of mean $w_t$ and fixed variance for continuous $\mathcal{X}$ or discrete distribution of probabilities $w_t$ for discrete $\mathcal{X}$.} is a distribution of parameters $w_t = f^\text{dec}_\phi(z_{1:t})$ outputted by a stacked SSM,
    \item A posterior distribution
    $q_\psi(z_{1:T}|x_{1:T})
    = \prod_{t=1}^T q_\psi(z_t|x_{1:t})
    = \prod_{t=1}^T \mathcal{D}(z_t|f_\psi^\text{enc}(x_{1:t}))$,
    where $\mathcal{D}(z_t|v_t)$ is a discrete distribution of probabilities $v_t = f^\text{enc}_\psi(x_{1:t})$ outputted by a stacked SSM.
\end{itemize}
The independence of the prior over all timesteps $z_t$, along with the conditional independence between $z_{\neq t}$ and $z_t$ given $x_{1:t}$ in $q_\psi$, and between $x_{\neq t}$ and $x_{t}$ given $z_{1:T}$ in $p_\phi$ enables the prior, posterior and generative models to be sampled in parallel.
Note that the discrete latent space requires the Gumbel reparametrization trick for computing $\nabla_\psi z_{1:T}$ when maximizing the ELBO \citep{jang2017categorical, maddison2016concrete}.

\subsection{Autoregressive Variational State Space Model} \label{subsec:partial}

In some applications, (e.g., language modeling) it is useful to learn a generative model of the distribution $p(x_{1:T}|x_{1:C})$ conditioned on a partial realization $x_{1:C}$.
Under the modeling assumptions of a trained dynamical VAE like the VSSM prior and generative models of \autoref{subsec:vssm}, we have,
\begin{align}
    p_\phi(x_{1:T}|x_{1:C}) = \int_{\mathcal{Z}^T} p_\phi(x_{1:T}|z_{1:T}) p_\phi(z_{1:T}|x_{1:C}) \diff z_{1:T},
\end{align}
where $p_\phi(x_{1:T}|z_{1:T})$ is our generative model,
while $p_\phi(z_{1:T}|x_{1:C})$ is the true partial posterior, from which we cannot sample for a given $x_{1:C}$.
We thus propose to learn an approximate partial posterior $q_\omega(z_{1:T}|x_{1:C})$ of the true partial posterior $p_\phi(z_{1:T}|x_{1:C})$, by exploiting samples $p(x_{1:T}|x_{1:C})$ from the dataset to construct samples of $p_\phi(z_{1:T}|x_{1:C})$ (see details \autoref{app:partial}).

The partial posterior $q_\omega(z_{1:T}|x_{1:C})$ is implemented with a stacked SSM,
where the input $x_{1:C}$ is padded with empty tokens: $\bar{x}_{1:T} = (x_{1:C}, \varnothing, \dots, \varnothing)$.
The autoregressive VSSM is a VSSM with,
\begin{itemize}
    \item A partial posterior distribution $\mathcal{D}(z_t|\bar{v}_t)$, where probabilities $\bar{v}_t = f^\text{par}_\omega(\bar{x}_{1:T})$ are the output of a stacked SSM, such that $q_\omega(z_{1:T}|x_{1:C}) = \prod_{t=1}^T q_\omega(z_t|x_{1:\min(C,t)}) = \prod_{t=1}^T \mathcal{D}(z_t|f^\text{par}_\omega(\bar{x}_{1:T}))$.
\end{itemize}
Note that the partial posterior distribution $q_\omega$ should ideally correspond to the prior when $\bar{x}_{1:T} = (\varnothing, \dots, \varnothing)$, and it will be used in practice for unconditional generation.

The autoregressive VSSM generates in parallel, possibly conditioned on a partial realization, and can resume generation, as illustrated on \autoref{fig:figure} (see detailed algorithms comparison in \autoref{app:sampling}).

\section{Experiments} \label{sec:experiment}

\begin{figure}[ht]
    \vspace{-0.5em}
    \centering
    \subfigure[\scriptsize MNIST valid log-likelihood.][b]{
        \label{fig:mnist_validation}
        \raisebox{-1em}{\includegraphics[width=0.28\textwidth]{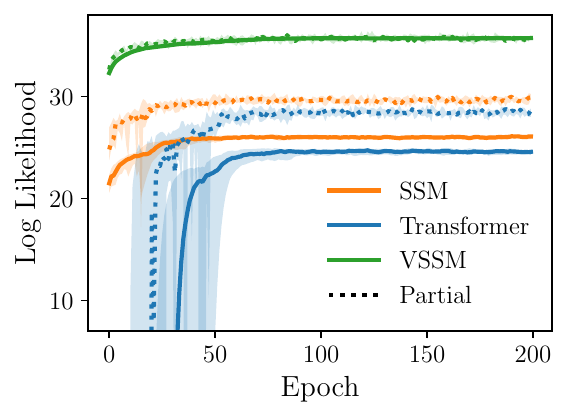}}
    }
    \subfigure[\scriptsize MNIST test statistics.][b]{
        \label{fig:mnist_test}
        \raisebox{-1em}{\includegraphics[width=0.28\textwidth]{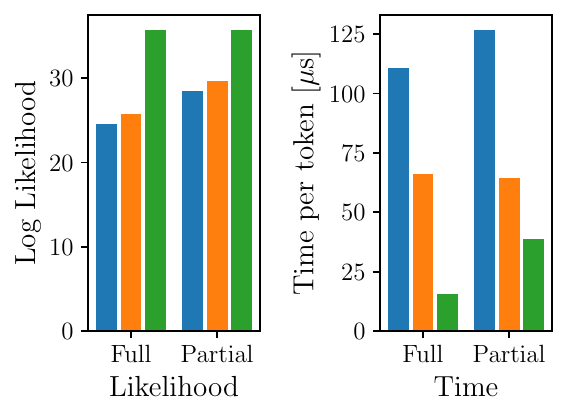}}
    }
    \subfigure[\scriptsize MNIST samples.][b]{
        \label{fig:mnist_samples}
        \raisebox{-1em}{\includegraphics[width=0.26\textwidth]{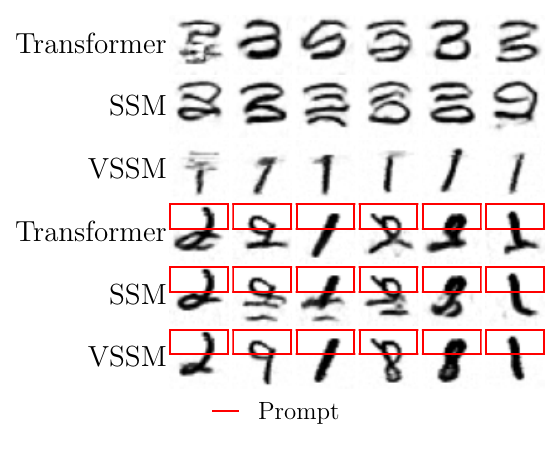}}
    }
    \subfigure[\scriptsize CIFAR valid log-likelihood.][b]{
        \label{fig:cifar_validation}
        \raisebox{-1em}{\includegraphics[width=0.28\textwidth]{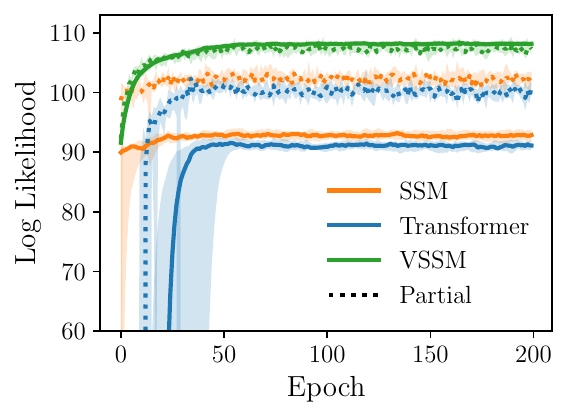}}
    }
    \subfigure[\scriptsize CIFAR test statistics.][b]{
        \label{fig:cifar_test}
        \raisebox{-1em}{\includegraphics[width=0.28\textwidth]{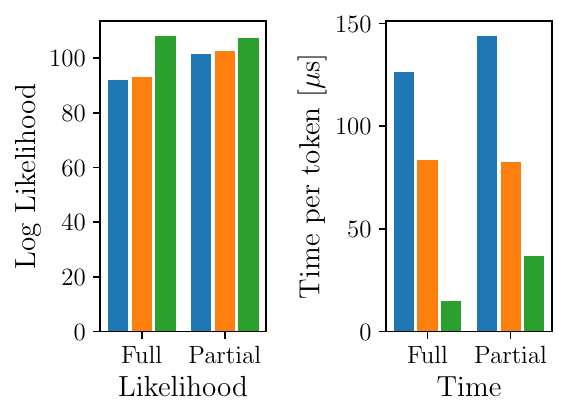}}
    }
    \subfigure[\scriptsize CIFAR samples.][b]{
        \label{fig:cifar_samples}
        \raisebox{-1em}{\includegraphics[width=0.26\textwidth]{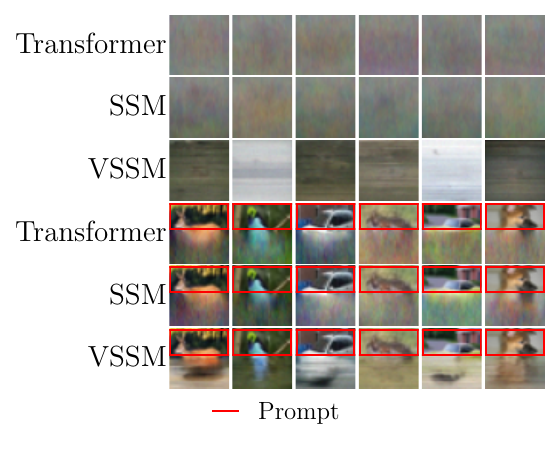}}
    }
    \caption{\small We report results over 5 runs of each model. Confidence intervals correspond to the minimum and maximum values observed. In \ref{fig:mnist_validation}, \ref{fig:cifar_validation}, we plot the median full and partial log-likelihood $\log p_\phi(x_{1:T})$ and $\log p_\phi(x_{C+1:T}\mid x_{1:C})$ on the validation set throughout training. In \ref{fig:mnist_test}, \ref{fig:cifar_test}, we report the average full and partial log-likelihood on the test set, along with mean execution times at generation, in both cases. In \ref{fig:mnist_samples}, \ref{fig:cifar_samples}, we report random qualitative examples for all models, for unconditioned sampling (first three rows) and conditioned on partial realizations (last three rows).}
    \vspace{-0.5em}
    \label{fig:results}
\end{figure}

In the following, we compare Transformer, SSM and VSSM on two toy sequence modeling tasks: MNIST, for which we consider 28-dimensional sequences of length 28, and CIFAR, for which we consider ($32 \times 3$)-dimensional sequences of length 32. Transformer and SSM both output the mean of a Gaussian distribution of fixed variance. For more details about model architectures, see \autoref{subsec:training}.
We report samples, generation times and likelihoods in \autoref{fig:results}, estimated by importance-sampling for the VSSM, see \autoref{subsec:evaluation}.
We report additional results in \autoref{subsec:results}.

\section{Conclusion} \label{sec:conclusion}

We introduce the VSSM, a dynamical VAE using SSMs as encoder and decoder.
Compared to other architectures, our model is the first one that can generate in parallel while being recurrent, which allows generation to be resumed.
Although tested on simple tasks, we show that it produces
decent results in only a fraction of the time.
The advantages of this architecture motivate further work to scale and improve performance on more challenging tasks such as language generation.

\acks{
    Gaspard Lambrechts gratefully acknowledges the financial support of the \emph{Wallonia-Brussels Federation} for his FRIA grant.
    Yann Claes gratefully acknowledges the financial support of the \emph{Walloon Region} under Grant No. 2010235 (ARIAC by Digital Wallonia 4.AI).
    Computational resources have been provided by the \emph{Consortium des Équipements de Calcul Intensif} (CÉCI), funded by the \emph{National Fund for Scientific Research} (F.R.S.-FNRS) under Grant No. 2502011 and by the \emph{Walloon Region}, including the Tier-1 supercomputer of the \emph{Wallonia-Brussels Federation}, infrastructure funded by the \emph{Walloon Region} under Grant No. 1117545.
}

\bibliography{references}
\newpage
\clearpage
\appendix

\section{Mathematical derivations} \label{app:derivations}

\subsection{Learning objective}
\label{app:learning_objective}

Thanks to Jensen's inequality, we can show for dynamical VAEs of \autoref{subsec:vae} that,
\begin{align}
    \log p_{\phi}(x_{1:T})
    &= \log \mathop{\mathbb{E}}_{p_\phi(z_{1:T})}
        p_\phi(x_{1:T}|z_{1:T}) \frac{q(z_{1:T}|x_{1:T})}{q(z_{1:T}|x_{1:T})}, \\
    &= \log \mathop{\mathbb{E}}_{q(z_{1:T}|x_{1:T})}
        \frac{p_\phi(x_{1:T}|z_{1:T}) p_\phi(z_{1:T})}{q(z_{1:T}|x_{1:T})}, \\
    &\geq \mathop{\mathbb{E}}_{q(z_{1:T}|x_{1:T})} \log\frac{p_\phi(x_{1:T}|z_{1:T}) p_\phi(z_{1:T})}{q(z_{1:T}|x_{1:T})}, \\
    &\geq \underbrace{
        \mathop{\mathbb{E}}_{q(z_{1:T}|x_{1:T})} \log p_\phi(x_{1:T}|z_{1:T}) - \operatorname{KL}(q(z_{1:T}|x_{1:T}) \parallel p_\phi(z_{1:T})),
    }_{
        \operatorname{ELBO}_\phi(x_{1:T})
    } \label{eq:elbo}
\end{align}

Note that the ELBO becomes tight when the inference model $q(z_{1:T}|x_{1:T})$ corresponds to the true posterior distribution $p_\phi(z_{1:T}|x_{1:T})$. Indeed,
\begin{align}
    \operatorname{ELBO}_\phi(x_{1:T})
        &= \mathop{\mathbb{E}}_{q(z_{1:T}|x_{1:T})}
            \log\frac{p_\phi(z_{1:T}|x_{1:T})p(x_{1:T})}{q(z_{1:T}|x_{1:T})}, \\
 & = \log p(x_{1:T}) - \operatorname{KL}(q(z_{1:T}|x_{1:T}) \parallel p_\phi(z_{1:T}|x_{1:T})),
\end{align}
and
$\operatorname{KL}(q(z_{1:T}|x_{1:T}) \parallel p_\phi(z_{1:T}|x_{1:T})) = 0$ if and only if $q(z_{1:T}|x_{1:T}) = p_\phi(z_{1:T}|x_{1:T})$ almost everywhere.
Hence, the dynamical VAE, composed of the prior $p_\phi(z_{1:T})$, generative model $p_\phi(x_{1:T}|z_{1:T})$ and inference model $q_\psi(z_{1:T}|x_{1:T})$ can be trained according to the objective function,
\begin{equation}
    \max_{\phi,\psi} \mathop{\mathbb{E}}_{p(x_{1:T})} \left[
        \mathop{\mathbb{E}}_{q_\psi(z_{1:T}|x_{1:T})} \left[
            \log p_\phi(x_{1:T}|z_{1:T})
        \right] - \operatorname{KL}(q_\psi(z_{1:T}|x_{1:T}) \parallel p_\phi(z_{1:T}))
    \right].
\end{equation}

\subsection{Approximate partial posterior} \label{app:partial}

To learn the approximate partial posterior $q_\omega(z_{1:T}|x_{1:C})$ of the true partial posterior $p_\phi(z_{1:T}|x_{1:C})$ introduced in \autoref{subsec:partial}, we propose to exploit samples of the true partial posterior.
Such samples are derived from samples $(x_{1:C}, x_{1:T})$, constructed from the dataset of sequences $x_{1:T}$ by taking random cuts $C \sim \mathcal{U}([0, T])$.
Indeed, these allow us to draw samples $(x_{1:C}, z_{1:T})$ such that $z_{1:T} \sim p_\phi(z_{1:T}|x_{1:C})$, as suggested by the decomposition,
\begin{align}
    p_\phi(z_{1:T}|x_{1:C}) = \int_{x_{1:T}} p_\phi(z_{1:T}|x_{1:T}) p(x_{1:T}|x_{1:C}) \diff x_{1:T},
\end{align}
where $p_\phi(z_{1:T}|x_{1:T})$ is the true posterior of this VAE, which we approximate by the variational posterior $q_\psi(z_{1:T}|x_{1:T})$ during the VSSM training.
The training objective for the approximate partial posterior $q_w(z_{1:T}|x_{1:C})$ is,

\label{app:partial_post_algo}
\begin{align}
    & \argmin_\omega \mathop{\mathbb{E}}_{p(x_{1:C})} \operatorname{KL}(p_\phi(z_{1:T}|x_{1:C}) \parallel q_\omega(z_{1:T}|x_{1:C})) \\
    =& \argmin_\omega \mathop{\mathbb{E}}_{p(x_{1:C})} \mathop{\mathbb{E}}_{p_\phi(z_{1:T}|x_{1:C})} [
        \log p_\phi(z_{1:T}|x_{1:C}) - \log q_\omega(z_{1:T}|x_{1:C})
    ] \\
    =& \argmax_\omega \mathop{\mathbb{E}}_{p(x_{1:C})} \mathop{\mathbb{E}}_{p_\phi(z_{1:T}|x_{1:C})} [
        \log q_\omega(z_{1:T}|x_{1:C})
    ] \\
    =& \argmax_\omega \mathop{\mathbb{E}}_{p(x_{1:C})} \mathop{\mathbb{E}}_{p(x_{1:T}|x_{1:C})} [
        \mathop{\mathbb{E}}_{p_\phi(z_{1:T}|x_{1:T})} [
            \log q_\omega(z_{1:T}|x_{1:C})
        ]
    ] \\
    \approx& \argmax_\omega \mathop{\mathbb{E}}_{p(x_{1:C})} \mathop{\mathbb{E}}_{p(x_{1:T}|x_{1:C})} [
        \mathop{\mathbb{E}}_{q_\psi(z_{1:T}|x_{1:T})} [
            \log q_\omega(z_{1:T}|x_{1:C})
        ]
    ] \label{eq:objective}
\end{align}

\section{Comparison of Autoregressive Generation} \label{app:sampling}

The VSSM sampling algorithm (\autoref{algo:vssm}) can be compared to the RNN (\autoref{algo:rnn}), SSM (\autoref{algo:ssm}), and Transformer (\autoref{algo:transformer}) algorithms.
We also provide an algorithm for the chunk sampling method proposed in \autoref{subsec:partial} in \autoref{algo:chunk}.

\begin{algorithm2e}[ht]
    \small
    \caption{VSSM sampling algorithm.} \label{algo:vssm}
    \DontPrintSemicolon
    \KwInput{Prompt $x_{1:C}$, length $T$.} \;
    Let $\bar{v}_{1:T} = f_\omega^\text{par}(k)$. \;
    Sample $z_t \sim \mathcal{D}(z_t|\bar{v}_t), \; t = 1, \dots, T$.\;
    Let $w_{1:T} = f_\phi^\text{dec}(z_{1:T})$. \;
    Sample $x_t \sim \mathcal{P}(x_t|w_t), \; t = C+1, \dots, T$. \;
    \Return $x_{1:T}$ \;
\end{algorithm2e}

\begin{algorithm2e}[ht]
    \small
    \caption{RNN sampling algorithm.} \label{algo:rnn}
    \DontPrintSemicolon
    \KwInput{Prompt $x_{1:C}$, length $T$.} \;
    Let $h_0 \leftarrow 0$. \;
    \For{$t \leftarrow 1, \dots, C$}{
        Let $h_t = f_\phi(x_t, h_{t-1})$. \;
    }
    \For{$t \leftarrow C+1, \dots, T$}{
        Let $w_{t} = g_\phi(h_{t-1})$. \;
        Sample $x_{t} \sim \mathcal{D}(x_{t}|w_{t})$. \;
        Let $h_{t} = f_\phi(x_t, h_{t-1})$. \;
    }
    \Return $x_{1:T}$ \;
\end{algorithm2e}

\begin{algorithm2e}[p]
    \small
    \caption{SSM sampling algorithm.} \label{algo:ssm}
    \DontPrintSemicolon
    \KwInput{Prompt $x_{1:C}$, length $T$.} \;
    Let $h_0 \leftarrow 0$. \;
    Let $h_C = f_\phi(x_{1:C}, h_0)$. \;
    \For{$t \leftarrow C+1, \dots, T$}{
        Let $w_{t} = g_\phi(h_{t-1})$. \;
        Sample $x_{t} \sim \mathcal{D}(x_{t}|w_{t})$. \;
        Let $h_{t} = f_\phi(x_t, h_{t-1})$. \;
    }
    \Return $x_{1:T}$ \;
\end{algorithm2e}

\begin{algorithm2e}[p]
    \small
    \caption{Transformer sampling algorithm.} \label{algo:transformer}
    \DontPrintSemicolon
    \KwInput{Prompt $x_{1:C}$, length $T$.} \;
    \For{$t \leftarrow C+1, \dots, T$}{
        Let $w_t = f_\phi(x_{1:t-1})$. \;
        Sample $x_t \sim \mathcal{D}(x_t|w_t)$.
    }
    \Return $x_{1:T}$ \;
\end{algorithm2e}

\begin{algorithm2e}[p]
    \small
    \caption{VSSM chunk sampling algorithm.} \label{algo:chunk}
    \DontPrintSemicolon
    \KwInput{Prompt $x_{1:C}$, length $T$, chunk size $W \in [1, T-C]$.} \;
    Let $(\bar{v}_{1:C}, h_C^\text{par}) = f_\omega^\text{par}(x_{1:C})$. \;
    Sample $z_t \sim \mathcal{D}(z_t|\bar{v}_t), \; t = 1, \dots, C$. \;
    Let $(w_{1:C}, h_C^\text{dec}) = f_\phi^\text{dec}(z_{1:C})$. \;
    \While{$C \leq T$}{
        Let $(\bar{v}_{C+1:C+W}, h_{C+W}^\text{par}) = f_\omega^\text{par}(\varnothing_{C+1:C+W}, h_C^\text{par})$. \;
        Sample $z_{t} \sim \mathcal{D}(z_{t}|\bar{v}_{t}), \; t = C+1, \dots, C+W$. \;
        Let $(w_{C+1:C+W}, h_{C+W}^\text{dec}) = f_\phi^\text{dec}(z_{C+1:C+W}, h_C^\text{dec})$. \;
        Sample $x_t \sim \mathcal{P}(x_t|w_t), \; t = C+1, \dots, C+W$. \;
        Update $C \leftarrow C + W$. \;
    }
    \Return $x_{1:T}$ \;
\end{algorithm2e}

\section{Additional details on experiments} \label{app:additional}

\subsection{Training details} \label{subsec:training}

We train all three architectures (Transformer, SSM, VSSM) with comparable sizes for $200$ epochs on the classical train set of the considered benchmarks (MNIST, CIFAR).
To prevent overfitting, we use $10\%$ of the train set for computing validation losses and likelihoods, as well as to select the final set of weights for evaluation on the test set and generating samples.
All three architectures use $4$ layers of dimension $1024$ (with a state size of $16$ for the SSM and VSSM, and with $8$ heads of dimension $1024 / 8 = 128$ for the Transformer).
We follow the attention block of GPT-2 for the Transformer, and the SSM block of Mamba for the SSM and VSSM architectures.
The SSM and Transformer architectures output the mean of a Gaussian distribution with fixed variance ($\sigma = 0.1$), and are trained to maximize the log-likelihood.
The VSSM generative model $p_\phi$ also outputs the mean of a Gaussian distribution with fixed variance ($\sigma = 0.1$), and is trained along with the posterior $q_\psi$ to maximize the ELBO \eqref{eq:elbo}.
Note that the temperature of the Gumbel softmax for computing $\nabla_\psi z_{1:T}$ was fixed to $1$.
The partial posterior $q_\omega$ is trained jointly with the encoder and decoder according to objective \eqref{eq:objective} and does not require to perform a subsequent training.
All learning rates have been selected using a grid search in ($\num{1e-2}, \num{5e-3}, \num{1e-3}, \num{5e-4}, \num{1e-4})$.

\subsection{Evaluation} \label{subsec:evaluation}

We evaluate the likelihood by sampling $K = 100$ latent variables \citep{burda2015importance} from the posterior, and reweighting by the prior (resp. partial posterior), in order to measure the likelihood (resp. the partial likelihood),
\begin{align}
    p_\phi(x_{1:T}) \approx \frac{1}{K} \sum_{k=1}^{K} p_\phi(x_{1:T}|z^k_{1:T}) \frac{p_\phi(z^k_{1:T})}{q_\psi(z^k_{1:T}|x_{1:T})}, \; z^k_{1:T} \sim q_\psi(z^k_{1:T}|x_{1:T}), \\
    p_\phi(x_{C+1:T}|x_{1:C}) \approx \frac{1}{K} \sum_{k=1}^{K} p_\phi(x_{C+1:T}|z^k_{1:T}) \frac{q_\omega(z^k_{1:T}|x_{1:C})}{q_\psi(z^k_{1:T}|x_{1:T})}, \; z^k_{1:T} \sim q_\psi(z^k_{1:T}|x_{1:T}).
\end{align}
This expression is known to be a lower bound on the likelihood in expectation, and it tends towards the true likelihood as $K$ grows to infinity.

\subsection{Additional results} \label{subsec:results}

\begin{figure}[ht]
    \vspace{-0.5em}
    \centering
    \subfigure[\scriptsize MNIST samples.][t]{
        \label{fig:mnist_additional}
        \includegraphics[width=0.98\textwidth]{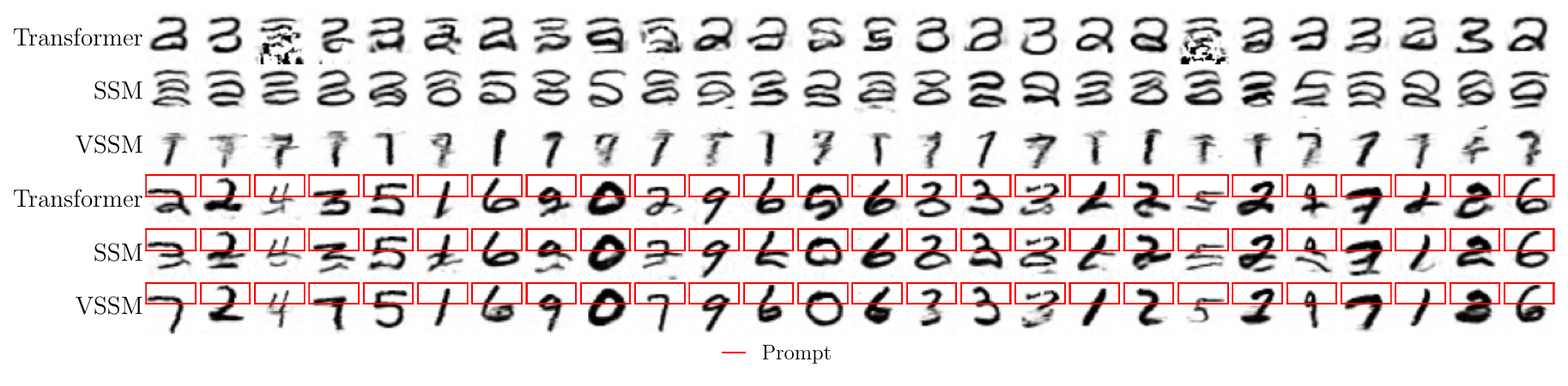}
    }
    \subfigure[\scriptsize CIFAR samples.][t]{
        \label{fig:cifar_additional}
        \includegraphics[width=0.98\textwidth]{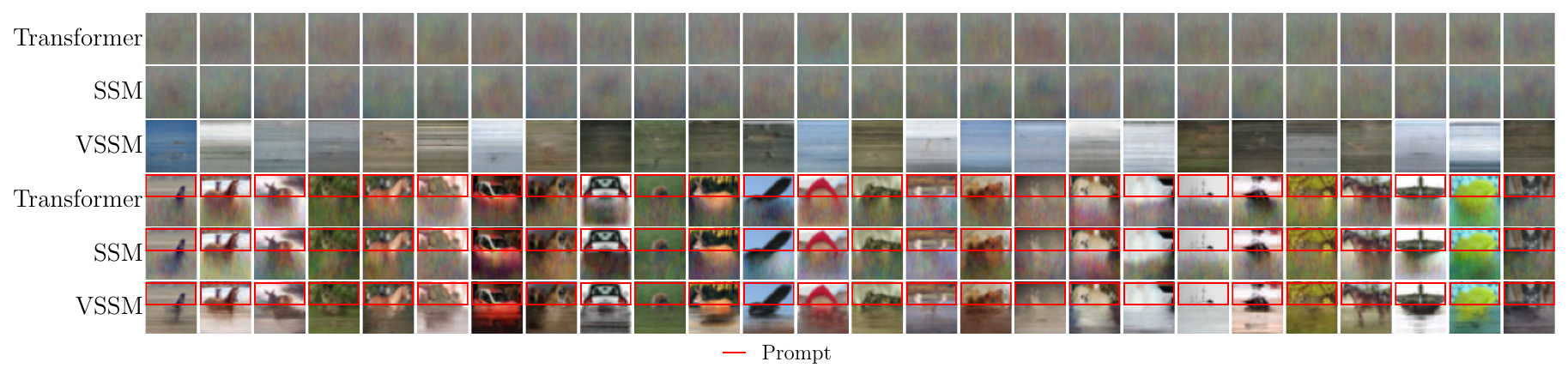}
    }
    \label{fig:additional}
    \caption{\small We report random qualitative examples for all models, for unconditioned sampling (first three rows) and conditioned on partial realizations (last three rows).}
    \vspace{-0.5em}
\end{figure}

We report additional samples from all models in \autoref{fig:mnist_additional} and \autoref{fig:cifar_additional}, sampled randomly and using random prompts from the test set.

\end{document}